\documentclass[letterpaper, 10 pt, conference]{ieeeconf}  

\IEEEoverridecommandlockouts                              

\usepackage{amsmath} 
\usepackage{amssymb}  
\usepackage[colorinlistoftodos]{todonotes}
\usepackage{hyperref}
\usepackage{multirow}
\usepackage{booktabs}
\usepackage[colorinlistoftodos]{todonotes}
\usepackage{siunitx}
\usepackage{url}

\usepackage[normalem]{ulem}

\usepackage{accents}

\newcommand{\vth}{\mbox{\boldmath $\theta$}}

\newcommand{\vlambda}{\mbox{\boldmath $\lambda$}}

\newcommand{\vnu}{\boldsymbol{ \nu}}

\newcommand{\vpi}{\mbox{\boldmath $\pi$}}

\newcommand{\vom}{\mbox{\boldmath $\omega$}}

\newcommand{\ve}{\mathbf e}
\newcommand{\vf}{\mathbf f}
\newcommand{\vg}{\mathbf g}
\newcommand{\vh}{\mathbf h}

\newcommand{\vn}{\mathbf n}

\newcommand{\vp}{\mathbf p}
\newcommand{\vq}{\mathbf q}
\newcommand{\vr}{\mathbf r}
\newcommand{\vs}{\mathbf s}

\newcommand{\vu}{\mathbf u}
\newcommand{\vv}{\mathbf v}

\newcommand{\vx}{\mathbf x}

\newcommand{\vA}{\mathbf A}
\newcommand{\vB}{\mathbf B}
\newcommand{\vC}{\mathbf C}
\newcommand{\vD}{\mathbf D}

\newcommand{\vF}{\mathbf F}

\newcommand{\vI}{\mathbf I}

\newcommand{\vK}{\mathbf K}

\newcommand{\vP}{\mathbf P}
\newcommand{\vQ}{\mathbf Q}
\newcommand{\vR}{\mathbf R}
\newcommand{\vS}{\mathbf S}
\newcommand{\vT}{\mathbf T}

\title{
Feedback MPC for Torque-Controlled Legged Robots
}

\author{Ruben Grandia$^1$, Farbod Farshidian$^1$, Ren\'{e} Ranftl$^2$, Marco Hutter$^1$
\thanks{This research was supported by Intel Network on Intelligent
Systems, the Swiss National Science Foundation through the National Centre of Competence in Research Robotics (NCCR Robotics), the European Union’s Horizon 2020 research and innovation programme under grant agreement No 780883. This work has been conducted as part of ANYmal Research, a community to advance legged robotics.}%
\thanks{$^1$ First, second, and last authors are with Robotic Systems Lab, ETH Zurich, Switzerland
        {\tt\footnotesize rgrandia@ethz.ch}
        {\tt\small}}%
\thanks{$^2$ Third author is with Intel Labs, Munich, Germany.
        {\tt\small }}%
}

\begin{document}

\maketitle

\begin{abstract}
The computational power of mobile robots is currently insufficient to achieve torque level whole-body Model Predictive Control (MPC) at the update rates required for complex dynamic systems such as legged robots. This problem is commonly circumvented by using a fast tracking controller to compensate for model errors between updates. In this work, we show that the feedback policy from a Differential Dynamic Programming (DDP) based MPC algorithm is a viable alternative to bridge the gap between the low MPC update rate and the actuation command rate. 
We propose to augment the DDP approach with a relaxed barrier function to address inequality constraints arising from the friction cone. A frequency-dependent cost function is used to reduce the sensitivity to high-frequency model errors and actuator bandwidth limits. We demonstrate that our approach can find stable locomotion policies for the torque-controlled quadruped, ANYmal, both in simulation and on hardware.
\end{abstract}

\section{Introduction}
Model Predictive Control (MPC) has gained broad interest in the robotics community as a tool for motion control of complex and dynamic systems. 
The ability to deal with nonlinearities and constraints has popularized the technique for many robotic applications, such as quadrotor control \cite{Alexis2011}, autonomous racing \cite{Liniger2015}, and legged locomotion~\cite{Tassa2012, Koenemann2015, Farshidian2017MPC}.

MPC strategies typically optimize an open-loop control sequence for a given cost function over a fixed time horizon.
The control sequence is then executed until a new control update is calculated based on the current state estimate.
While this strategy assumes that the model is exact and that there are no external disturbances,
the repeated optimization provides a feedback mechanism that can correct for modeling errors provided that the control loop can be executed at a sufficiently high rate.
However, for high dimensional systems such as legged robots and due to the computational restrictions of mobile platforms, the achievable update rate of the MPC loop is insufficient to effectively deal with model uncertainty and external disturbances. 

As a remedy, a separately designed, light-weight motion tracker is often used in practice \cite{murray2009}. 
The motion tracker runs at a higher rate than the MPC loop and provides feedback correction to the control sequence that was designed by the MPC. 
For complex systems such as legged robots it is a challenging task to design a controller that tracks arbitrary motions while satisfying the many constraints arising from the locomotion task.
The fundamental problem is that such motion trackers do not look ahead in the horizon and therefore cannot anticipate changes in contact configuration. 
As an alternative, projected (time-varying) Linear Quadratic Regulators (LQR) have been proposed as a framework to automatically design feedback controllers around a given reference trajectory \cite{posa2016, mason2016}. However, the stabilizing feedback policy is always designed in a secondary stage and often with a different objective function than the one used for computing the optimal trajectories, which leads to inconsistency between the feedback policy and the MPC trajectories. 
\begin{figure}[!t]
    \centering
    \includegraphics[trim=0mm 20mm 10mm 30mm, clip,width=1.0\columnwidth]{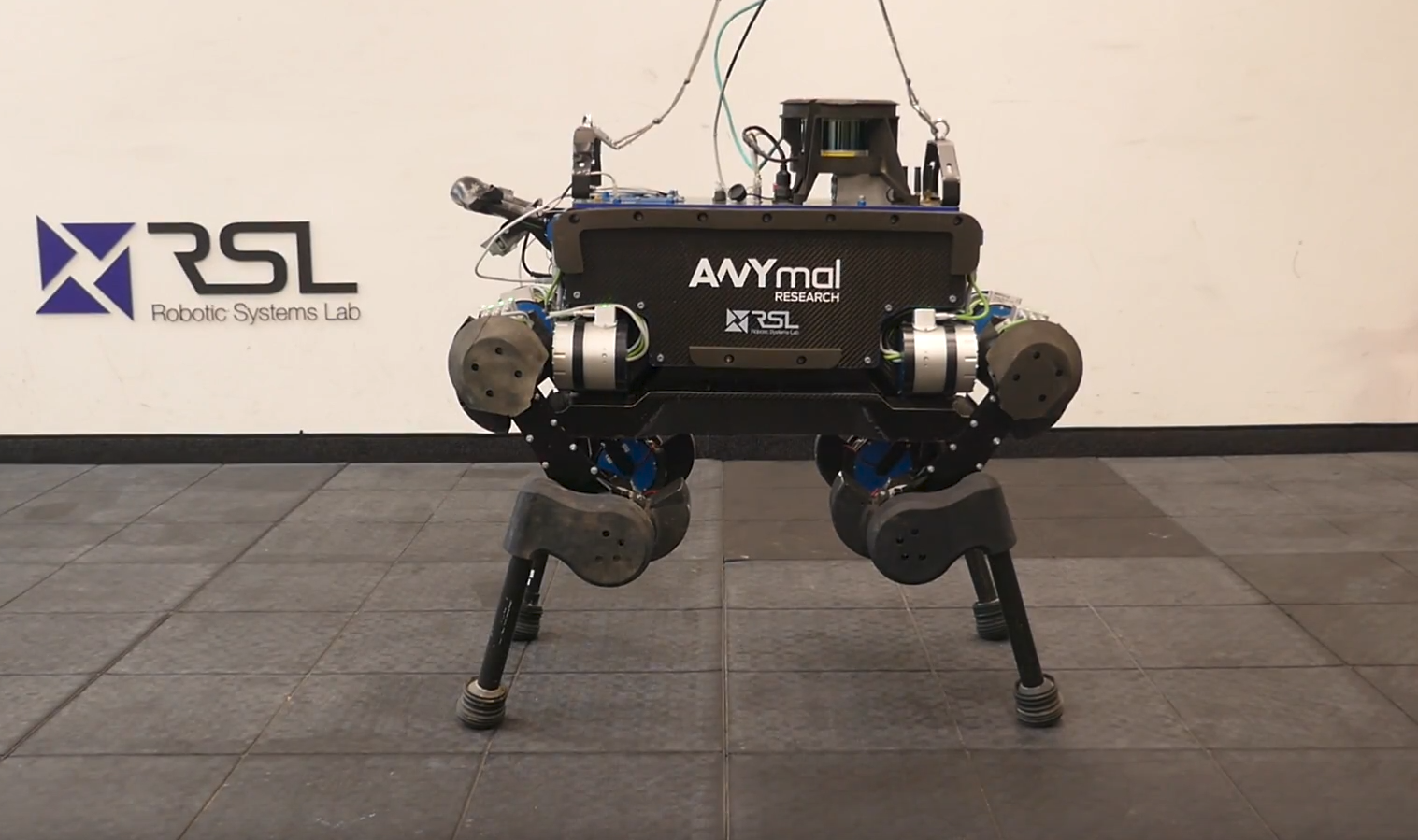}
    \vspace*{-5mm}
    \caption{ANYmal, the torque controlled-legged robot used in this work.}
    \label{fig:ANYmal}
    \vspace{-5mm}
\end{figure}


In this work, we propose a feedback MPC approach for motion control of a legged system and show that the optimized feedback policy can directly be deployed on hardware. We achieve stable locomotion under a very low update rate (15 Hz), and the optimized feedback policy removes the need for a separate motion controller. Furthermore, the modification of the control inputs is consistent with the MPC and thus produces a continuous signal across update instances. 

To be able to directly apply the feedback strategy on a legged system, the optimized policy needs to respect all the constraints of the locomotion task such as friction and unilateral constraints of contact forces. 
To achieve this, we propose to extend the SLQ (Sequential Linear Quadratic) algorithm \cite{Farshidian2017SLQ}. We use SLQ in a real-time iteration MPC scheme \cite{diehl2005} where the algorithm optimizes a constrained feedback policy, $\vpi(\vx,t): \mathbb{X} \times \mathbb{R}^+ \!\to\! \mathbb{U}$,
\begin{align}
\label{eq:feedback_policy}
\vpi(\vx,t) = \vu^*(t) + \vK(t) \left( \vx - \vx^*(t) \right) ,
\end{align}
where ${\vu^*(t) \in \mathbb{U}}$ and ${\vx^*(t) \in \mathbb{X}}$ are locally optimal input and state trajectories. $\vK(t)$ is a time-varying LQR gain matrix which maps the state deviation from $\vx^*$ to an admissible control correction. We extend the algorithm to problems with inequality constraints using a barrier function method to accurately handle the constraints arising from the friction cone. We further use the frequency-aware MPC approach introduced in \cite{Grandia2019MPC} to render the resulting feedback policy robust to the bandwidth limitations imposed by real actuators.

We perform experiments in simulation and on a real legged system (Fig. \ref{fig:ANYmal}) and demonstrate that our approach is able to find robust and stable locomotion polices at MPC update rates as low as 15 Hz, which facilitates onboard execution on mobile platforms with limited computational power.



\subsection{Related Work}
\label{sec:related_work}
Methods to incorporate robustness explicitly into the MPC methodology have been previously studied in the literature~\cite{Mayne2000}. Min-max MPC \cite{Bemporad1999}, for example, optimizes an open-loop control sequence for the worst-case disturbance inside a predefined set.
While this formulation appears attractive, it can be overly conservative due to its inability to include the notion of feedback that is inherently present in the receding-horizon implementation of the control \cite{Lee1997}. 

Min-max Feedback MPC was proposed to address this shortcoming by planning over a state-dependent control policy instead of an open-loop feedforward sequence \cite{Scokaert1998}. Unfortunately optimizing the feedback policy for all possible disturbance realizations does not yet scale to the problem dimensions encountered in legged robotics. However, even without considering disturbances, optimizing over the feedback policy has an additional advantage. When the update rate of the MPC loop is low, the feedback policy can provide local correction to the deviation of the real platform from the optimal trajectories. 
Exploiting this additional aspect of feedback MPC has not yet been fully explored in robotic applications that are subject to path constraints. 

The feedback policy that minimizes a cost function for a given dynamical system and path constraints can be computed using the Hamilton-Jacobi-Bellman (HJB) equation \cite{Bertsekas1995}.
While directly solving this equation for high dimensional systems is prohibitively complex, a variant of the dynamic programming approach known as Differential Dynamic Programming (DDP) \cite{mayne1966DDP} has proven to be a powerful tool in many practical applications. 
The SLQ method that we use in this work is a DDP-based approach which uses a Gauss-Newton approximation. Consequently, it only considers the linearized dynamics instead of a second-order approximation. 

Although using the LQR gains derived from a DDP-based approach directly for motion tracking generates promising results in simulation, it dramatically fails on real hardware. 
This phenomenon has been reported before in other real-world applications of LQR on torque-controlled robots \cite{mason2016,mason2014}.
Focchi et al. \cite{focchi2016} have shown that instability can occur if the limitations of the low-level torque controller are neglected in the high-level control design. 
They have argued that the bandwidth of the low-level controller inversely relates to the achievable impedance of the high-level controller. 
To this end, to apply the SLQ feedback policy on hardware, we need to encode these bandwidth limitations in our optimization problem. 
In this work, we use the frequency-aware MPC approach introduced in \cite{Grandia2019MPC}. 
This MPC formulation penalizes control actions in the frequency domain and automatically finds a trade-off between the bandwidth limitation of actuators and the stiffness of the high-level feedback policy.

\subsection{Contributions}

We propose a whole-body MPC approach for legged robots,
where the actuation commands are computed directly based on the MPC feedback policy. 
Specifically we present the following contributions which we empirically validate on the ANYmal platform (Fig.~\ref{fig:ANYmal}) in  simulation and on real hardware:

\begin{itemize}
    \item We propose to apply feedback MPC for whole-body control of a legged system. To the best of our knowledge, this is the first time that such a control scheme is applied on hardware for motion control of legged robots.
    \item The SLQ algorithm is extended to include inequality constraints through a barrier function method, which allows us to formulate friction cone constraints.
    \item We show that our feedback MPC algorithm directly designs constraint-satisfactory LQR gains without additional computational cost. 
    \item A frequency domain design approach is used to incorporate actuation bandwidth limits in the MPC formulation to avoid rendering stiff gains. Thus, the feedback gains can be directly applied to the robot.  
    \item  We show that the feedback MPC algorithm is capable of bridging the gap between low update-rate MPC and high rate execution of torque commands using only an onboard computer with moderate computational power.
\end{itemize}

\section{Method}

\subsection{Problem Definition}
Consider the following nonlinear optimal control problem with cost functional 
\begin{equation}
\min_{u(\cdot)}   \Phi(\vx(T)) + \int_{0}^{T} L(\vx(t),\vu(t), t) \, dt, 
 \label{eq:mpc_cost}
\end{equation}
where $\vx(t)$ is the state and $\vu(t)$ is the input at time $t$. $L(\cdot)$ is a time-varying running cost, and $\Phi(\cdot)$ is the cost at the terminal state $\vx(T)$.
Our goal is to find an input trajectory $\vu(\cdot)$ that minimizes this cost subject to the following system dynamics, initial condition, and general equality and inequality constraints:
\begin{align}
& \dot{\vx} =  f(\vx, \vu, t)  \label{eq:mpc_dynamics} \\
& \vx(0) = \vx_0  \label{eq:mpc_initial} \\
& \vg_1(\vx,\vu, t) =  0  \label{eq:mpc_eqconstraint} \\
& \vg_2(\vx, t) =  0  \label{eq:mpc_eqconstraint_state} \\
& \vh(\vx,\vu, t) \geq  0 \label{eq:mpc_inequality}.
\end{align}

The feedback policy which minimizes this problem can be calculated using a DDP-based method. 
A variant of this method for continuous-time systems known as SLQ is introduced in \cite{Farshidian2017SLQ}, where it solves the above optimization problem in the absence of the inequality constraints in equation~\eqref{eq:mpc_inequality}. 
This method computes a time-varying, state-affine control policy based on a quadratic approximation of the optimal value function in an iterative process.
The SLQ approach uses a Lagrangian method to enforce the state-input equality constraints in \eqref{eq:mpc_eqconstraint}. The pure state constraints in \eqref{eq:mpc_eqconstraint_state} are handled by a penalty method. 

In SLQ, the simulation (forward pass) and the optimization (backward pass) iterations alternate. Once the backward pass is completed, a forward pass computes a new trajectory based on the improved feedback policy.
The local, Linear Quadratic (LQ), approximation of the nonlinear optimal control problem is constructed after each forward pass. 
The LQ model permits an efficient solution of the approximate problem by solving the Riccati differential equation. 
The feedback policy is then updated with an appropriate linesearch procedure in the direction of the LQ problem's solution. 

We follow the same SLQ approach and extend the method with inequality constraints through a relaxed barrier function approach. 

\subsection{Relaxed Barrier Functions}
Using a barrier function is a well know technique to absorb inequality constraint into the cost function. For each constraint in a given set of $N_{in}$ inequality constraints, a barrier term $B(h)$ is added to the cost
\begin{equation}
    \hat{L}(\vx, \vu, t) = L(\vx, \vu, t) + \mu \sum_{i=1}^{N_{in}} B(h_i(\vx, \vu, t)).
    \label{eq:cost_plus_barrier}
\end{equation}

A widely used barrier function is the logarithmic barrier used in interior-point methods. The optimal solution is approached by letting $\mu \rightarrow 0$ over successive iterations. However, a downside of the log-barrier is that it is only defined over the feasible space, and evaluates to infinity outside. Due to the rollout mechanism in the SLQ approach, one cannot ensure that successive iterations remain inside the feasible region at all time. Furthermore, the Hessian of the log-barrier goes to infinity as one approaches the constraint boundary, which results in an ill-conditioned LQ approximation. 

The relaxed barrier functions previously proposed for MPC problems addresses both these issues \cite{Feller2017} and is therefore particularly suitable for the SLQ approach. This barrier function is defined as a log-barrier function on the interior of the feasible space, and switched to a different function at a distance $\delta$ from the constraint boundary.
\begin{equation}
        B(h)= 
\begin{cases}
    - \ln(h) , & h \geq \delta, \\
    \beta(h; \delta),  & h < \delta.
\end{cases}
\label{eq:relaxed_barrier}
\end{equation}
We use the quadratic extension proposed in \cite{Hauser2006}:
\begin{equation}
\beta(h; \delta) = \frac{1}{2}\left(\left(\frac{h - 2\delta}{\delta}\right)^2 - 1 \right) -\ln(\delta).
\end{equation}

\begin{figure}[!t]
    \centering
    \includegraphics[width=0.9\columnwidth]{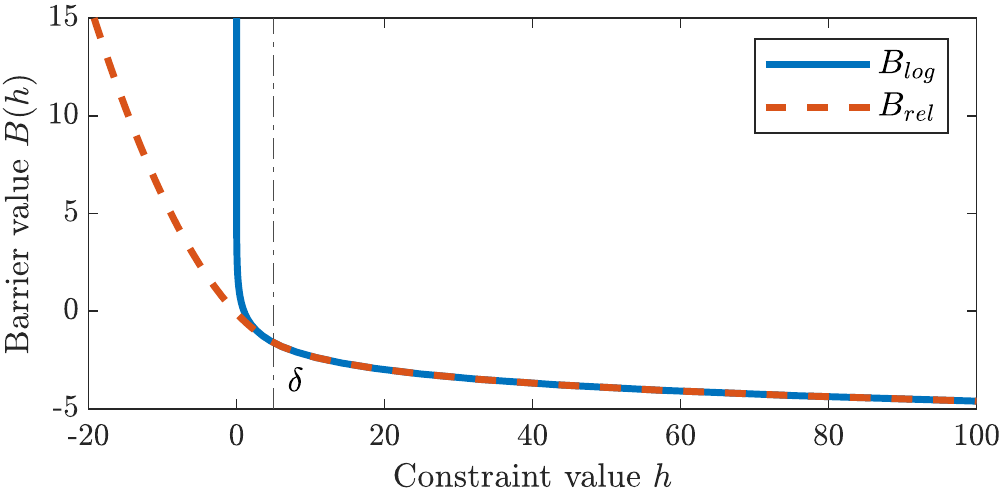}
    \vspace*{-2mm}
    \caption{Comparison of the log-barrier function $B_{log} = -\ln(h)$ and relaxed barrier function $B_{rel}$ as defined in \eqref{eq:relaxed_barrier} with $\delta = 5$.}
    \label{fig:barrierComparison}
    \vspace{-5mm}
\end{figure}

The relaxed barrier function, which is continuous and twice differentiable, is plotted as a function of the constraint value in Fig.~\ref{fig:barrierComparison}. The quadratic extension puts an upper bound to the curvature of the barrier function, which prevents ill-conditioning of the LQ approximation. Note that by letting $\delta \rightarrow 0$, the standard logarithmic barrier is retrieved. Furthermore, it has been shown that the optimal solution can be obtained for a nonzero value of $\delta$ \cite{Aguiar2017}, when the gradient of the penalty term is larger than the Lagrange multiplier of the associated constraint. Optimization with the relaxed barrier function can thus be interpreted as an augmented Lagrangian approach when $h < \delta$ and as a log-barrier method for $h \geq \delta$. 

\subsection{LQ Approximation}
With the inequality constraints embedded in the cost function, we obtain the following linearization of the system dynamics in \eqref{eq:mpc_dynamics} and state-inputs constraints in \eqref{eq:mpc_eqconstraint} for a given state trajectory $\vx_{k-1}(t)$ and input trajectory $\vu_{k-1}(t)$:
\begin{align}
    &\delta\dot{\vx}˙ = \vA \delta \vx + \vB \delta \vu, \\
    & \vC \delta\vx + \vD \delta\vu + \ve = 0,
\end{align}
where $\delta \vx = \vx(t) - \vx_{k-1}(t)$, and $\delta \vu = \vu(t) - \vu_{k-1}(t)$ are deviations from the previous iteration, around which the LQ approximation is made. 
Note that the time-dependency of the matrices was dropped to shorten the notation. 
The quadratic approximation of the cost in \eqref{eq:cost_plus_barrier} is given by
\begin{align}
 \Phi(\vx(T)) &\approx  q_f + \vq_f^T \delta\vx + \frac{1}{2} \delta\vx^T \vQ_f^T \delta\vx, \\
     \hat{L}(\vx,\vu, t) &\approx q_L(t) + \vq_L^T \delta\vx + \vr^T \delta\vu + \notag \\ &\frac{1}{2}\delta\vx^T \vQ_L \delta\vx + \frac{1}{2}\delta\vu^T \vR \delta\vu + \delta\vu^T \vP \delta\vx, \label{eq:quadraticCostApprox}
\end{align}
which requires access to the second-order approximation of the barrier term and inequality constraints.




With the optimal control problem reduced to an equality constrained LQ approximation, the constrained Riccati backward pass in \cite{Farshidian2017SLQ} yields the quadratic value function $V(\vx, t) = \frac{1}{2} \vx^T \vS(t) \vx + \vs^T(t) \vx + s(t)$. This value function induces the optimal feedback policy in equation~\eqref{eq:feedback_policy} with feedback gains computed as
\begin{align}
    &\vK(t) = \left(\vI - \vD^\dagger \vD\right)\vR^{-1}\left(\vP^T + \vB^T\vS\right) + \vD^\dagger\vC, \label{eq:feedbackmatrix_backwardpass}
\end{align}
where $\vI$ is the identity matrix and $\vD^\dagger$ is the right pseudo-inverse of the full row rank matrix $\vD$. Notice how the feedback gains ensure that the equality constraints are satisfied by projecting the first term to the nullspace of the constraints, and by adding the term $\vD^\dagger\vC$ to satisfy the constraint when the state deviates from the plan.

\subsection{Frequency Shaping}
As discussed in Section~\ref{sec:related_work}, it has been proven difficult to use feedback gains from an LQR design on a torque-controlled robot. We propose to use the frequency-dependent cost function introduced in our previous work \cite{Grandia2019MPC}, which was used to render the feedforward solution robust to high frequency disturbances. In this work, we show that it has a similar effect on the feedback structure. We briefly summarize how the problem is adapted and refer to \cite{Grandia2019MPC} for further details.

A frequency-dependent cost on the inputs can be introduced by evaluating the cost function on auxiliary inputs $\vnu$. The auxiliary inputs $\vnu$ are defined by frequency-dependent shaping functions $r_i(\omega)$ applied to the system inputs such that
\begin{align}
    \hat{\nu}_i(\omega) & = r_i(\omega) \hat{u}_i(\omega),
\end{align}
where $i$ denotes elements associated to individual inputs, $\omega$ is the signal frequency in \SI{}{\radian\per\second}, and $\hat{\nu}_i(\omega)$ and $\hat{u}_i(\omega)$ are the Fourier transform of the auxiliary input and system input respectively.
Following our previous work, we use high pass filters to achieve increased costs at higher input frequencies:
\begin{align} 
r_i(\omega) & = \frac{1 + \beta_i j\omega}{1 + \alpha_i j\omega}, \qquad \beta_i>\alpha_i.
    \label{eq:individualweightingfunction}
\end{align}
The transfer function $\vs(\omega) = \vr^{-1}(\omega)$, with state space realization ${(\vA_s, \vB_s, \vC_s, \vD_s)}$, is constructed such that ${\hat{u}_i(\omega) = s_i(\omega) \hat{\nu}_i(\omega)}$. The original system is augmented with an additional filter state, $\vx_s$, such that $\tilde{\vx} = [\vx^T,  \vx_s^T]^T$, and optimization is performed w.r.t the auxiliary inputs $\vnu$. The augmented system dynamics and state-input constraints are defined as
\begin{equation}
    \begin{bmatrix}
\vf(\vx, \vC_s  \vx_{s} + \vD_s \vnu)  \\
 \vA_s  \vx_{s} + \vB_s \vnu
\end{bmatrix}, \quad 
\begin{matrix}
\vg_1(\vx, \vC_s  \vx_{s} + \vD_s \vnu, t)  = 0 \\
\vh(\vx, \vC_s  \vx_{s} + \vD_s \vnu, t) \geq  0
\end{matrix}.
\end{equation}

The feedback policy obtained from this augmented system is of the form 
 \begin{equation}
    \vnu(\vx, t) = \vnu^*(t) +  \begin{bmatrix}
  \vK_{\vnu,\vx}(t) & \vK_{\vnu,\vx_s}(t)  
 \end{bmatrix}     
 \begin{bmatrix}
  \vx - \vx^*(t) \\
  \vx_s - \vx_s^*(t)
 \end{bmatrix} \nonumber.
\end{equation}

After optimization, the original input is retrieved by substituting this policy into the output function of the filter, $\vu =  \vC_s \vx_{s} + \vD_s \vnu$, resulting in the complete feedback policy
\begin{align}
 & \begin{bmatrix}
  \vu(\tilde{\vx}, t) \\
  \vnu(\tilde{\vx}, t) 
 \end{bmatrix} =     \begin{bmatrix}
 \vC_s \vx_s^*(t) + \vD_s \vnu^*(t)  \notag \\
  \vnu^*(t) 
 \end{bmatrix} +  \\
   & \quad  \begin{bmatrix}
  \vD_s \vK_{\vnu,\vx}(t) & \vD_s \vK_{\vnu,\vx_s}(t) + \vC_s  \\
  \vK_{\vnu,\vx}(t) & \vK_{\vnu,\vx_s}(t)  
 \end{bmatrix}     
 \begin{bmatrix}
  \vx - \vx^*(t) \\
  \vx_s - \vx_s^*(t)
 \end{bmatrix}. \label{eq:freq_feedback}
\end{align}

\section{Implementation}

We apply our approach to the kinodynamic model of a quadruped robot, which describes the dynamics of a single free-floating body along with the kinematics for each leg. The Equations of Motion (EoM) are given by
\begin{align*}
&\left\{ 
\begin{array}{ll}
		\dot{\vth} = \vT(\vth) \vom \\
		\dot{\vp}  = _W\!\vR_B(\vth) \, \vv \\
		\dot{\vom} = \vI^{-1} \left(  -\vom \times \vI \vom + \sum_{i=1}^4{ {\vr_{EE}}_j(\vq) \times {\vlambda_{EE}}_j} \right)\\
		\dot{\vv}  = \vg(\vth) + \frac{1}{m} \sum_{i=1}^4{\vlambda_{EE}}_j \\
		\dot{\vq} = \vu_J,
\end{array}	 
\right. \notag
\end{align*}
where $_W\!\vR_B$ and $\vT$ are the rotation matrix of the base with respect the global frame and 
the transformation matrix from angular velocities in the base frame to the Euler angles derivatives in the global frame.
$\vg$ is the gravitational acceleration in body frame, $\vI$ and $m$ are the moment of inertia about the CoM and the total mass respectively. The inertia is assumed to be constant and taken at the default configuration of the robot. ${\vr_{EE}}_j$ is the position of the foot $j$ with respect to CoM. 
$\vth$ is the orientation of the base in Euler angles, $\vp$ is the position of the CoM in world frame, $\vom$ is the angular rate, and $\vv$ is the linear velocity of the CoM. $\vq$~is the vector of twelve joint positions. The inputs of the model are the joint velocity commands $\vu_J$ and end-effector contact forces ${\vlambda_{EE}}_j$ in the body frame. 

\subsection{Equality Constraints}
The equality constraints depend on the mode of each leg at a certain point in time. We assume that the mode sequence is a predefined function of time. The resulting mode-depended constraints are
\begin{align*}
&\left\{ 
\begin{array}{ll}
		{\vv_{EE}}_j = \mathbf{0}, \quad &\text{if $i$ is a stance leg}, \\
		{\vv_{EE}}_j \cdot \hat{\vn} = c(t), \quad  {\vlambda_{EE}}_j = \mathbf{0}, \quad  &\text{if $i$ is a swing leg},
\end{array}
\right.
\end{align*}
where ${\vv_{EE}}_j$ is the end-effector velocity in world frame.  
These constraints ensure that a stance leg remains on the ground and a swing leg follows the predefined curve $c(t)$ in the direction of the local surface normal $\hat{\vn}$ to avoid foot scuffing. 
Furthermore, the constraints enforce zero contact force at swing legs. 

\subsection{Inequality Constraints}
Our proposed relaxed barrier method allows to model the friction cone without the commonly used polytope approximation. The cone constraint for each end-effector,
\begin{equation}
    {\vlambda_{EE}}_j \in \mathcal{C}(\hat{\vn}, \mu_c),
\end{equation}
is defined by the surface normal and friction coefficient ${\mu_c = 0.7}$. After projecting the contact forces to the local frame of the surface, a canonical second-order cone constraint is found in terms of local contact forces ${\vF = [F_x, F_y, F_z]}$. An effective cone constraint used in conjunction with barrier methods \cite{lobo1998applications} is given by
\begin{equation}
    h_{cone} = \mu_c F_z - \sqrt{F_x^2 + F_y^2} \geq 0.
    \label{eq:cone}
\end{equation}
However, the gradient of this constraint is not defined at $\vF=0$, which causes numerical issues close to the origin. While for interior point methods this problem can be solved by using the squared constraint, $\mu_c^2 F_z^2 - F_x^2 + F_y^2 \geq 0$, 
this strategy does not work well together with the relaxed barrier function due to the saddle point it introduces at the origin. Since the relaxed barrier function allows infeasible iterates, the solutions can cross the origin and end up in the negative reflection of the cone, which became a feasible region through the squaring operation. We, therefore, use the perturbed cone
\begin{equation}
    h_{cone, \epsilon} = \mu_c F_z - \sqrt{F_x^2 + F_y^2 + \epsilon^2} \geq 0,
\end{equation}
which is differentiable at the origin, remains infeasible for any negative $F_z$, and is a conservative lower bound for the original cone \eqref{eq:cone}. It therefore holds that
\begin{equation}
    h_{cone, \epsilon} \geq 0 \implies h_{cone}  \geq 0.
\end{equation}

\begin{figure}[!t]
    \centering
    \includegraphics[width=0.9\columnwidth]{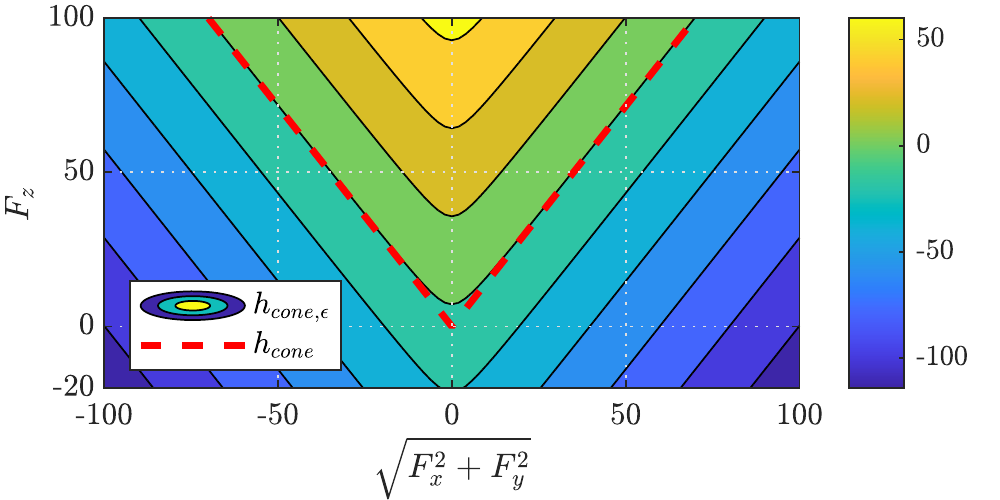}
    \vspace*{-2mm}
    \caption{Comparison of the friction cone constraint $h_{cone}$ and the perturbed cone $h_{cone, \epsilon}$ for $\epsilon = 5$, $\mu_c = 0.7$. The contour of $h_{cone, \epsilon}$ is shown together with the zero crossing of $h_{cone}$.}
    \label{fig:coneConstraint}
    \vspace{-5mm}
\end{figure}

In Fig.~\ref{fig:coneConstraint} the level sets of this constraint are compared to the original cone. It can be seen that the constraint is convex and the zero crossing of $h_{cone, \epsilon}$ is strictly inside the feasible region. 

\subsection{Torque Computation}
The control inputs $\vu$ consist of contact forces and joint velocities. These commands have to be translated to torques. When using only the feedforward trajectories, desired accelerations and contact forces are extracted and tracked by a hierarchical inverse dynamics controller \cite{bellicoso2016perception}. When using the feedback policy, we forward simulate the system under the feedback policy for a short time and extract desired accelerations from this rollout. The inverse dynamics is then only used to convert the desired accelerations into torques, without adding additional feedback. This inverse dynamics controller is evaluated at \SI{400}{\hertz}, while the SLQ-MPC algorithm runs asynchronously on a second onboard Intel i7-4600U@2.1GHz dual core processor.

Finally, each individual motor has a local, embedded, control loop. For the stance legs a torque controller is used, and for the swing legs the motors take the commanded torque as a feedforward term and close the loop over the desired position and joint velocities. 

\section{Results}
We first show the qualitative differences between a feedback policy and a feedforward policy when operating under low update rates and disturbances. Then, the influence of the relaxed barrier function cone constraint on the planned feedback gains is shown. We also examine the structure in the obtained feedback matrices and compare the difference between those obtained with frequency shaping and those without. Finally, we show that the proposed elements, when taken together, lead to a method that can be successfully executed on the onboard hardware of a torque-controlled robot.

We use a diagonal cost on the state and control inputs for all experiments. When frequency shaping is used, we set $\alpha_i = 0.01$, $\beta_i = 0.2$ for the contact force inputs and $\alpha_i = 0.01$, $\beta_i = 0.1$ for the joint velocity inputs in \eqref{eq:individualweightingfunction}.

\subsection{Feedback MPC}
We first investigate the effect of low update frequencies and the use of feedback policies from the SLQ-MPC in simulation. We introduce model errors to show the performance of the different strategies. The mass of the control model is increased by \SI{10}{\percent} with respect to the simulation model. Each MPC controller is brought to a stable trot gait and commanded to move \SI{1}{\meter} forward. 

\begin{figure}[!t]
    \centering
    \includegraphics[width=1.0\columnwidth]{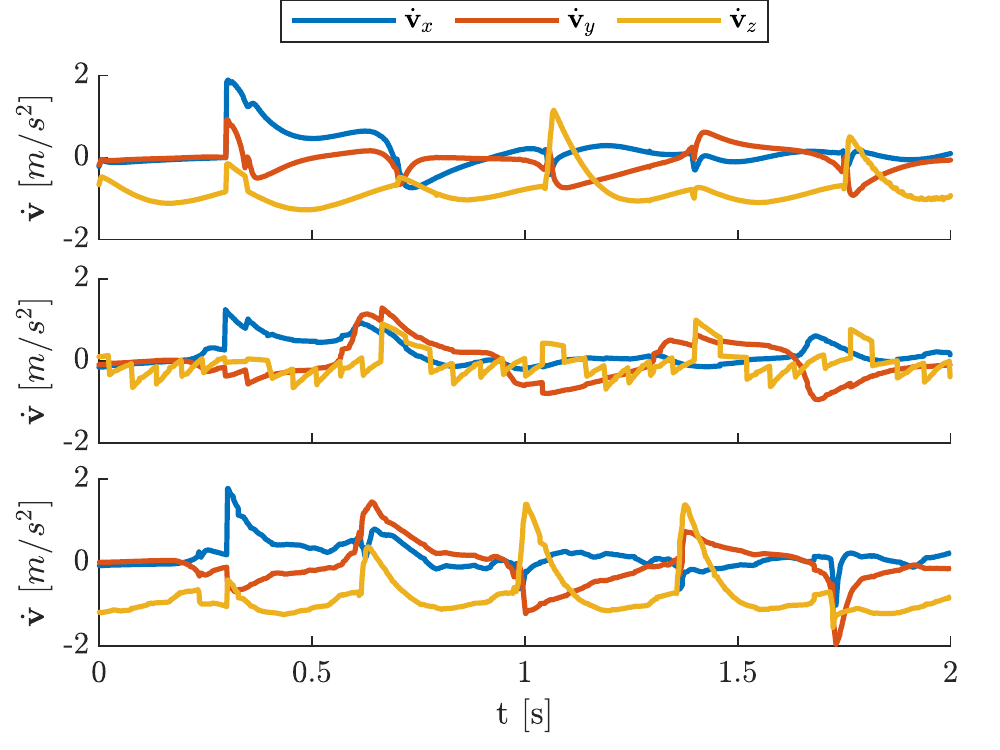}
    \vspace*{-7mm}
    \caption{Desired linear acceleration for the center of mass, as commanded by the MPC. In the top two plots a feedforward MPC strategy is used with update rates of \SI{400}{\hertz} and \SI{20}{\hertz} respectively. In the bottom plot, feedback MPC is used with an update rate of \SI{20}{\hertz}. The trajectories are for a trotting gait with a command to move \SI{1}{\meter} ahead send at \SI{0.3}{\second}}
    \label{fig:trotAcceleration}
    \vspace{-3mm}
\end{figure}

First, feedforward MPC is used with an update rate equal to the control frequency of \SI{400}{\hertz}. Every control loop thus has access to the optimal solution from the current state. The resulting desired linear accelerations are shown in the top of  Fig.~\ref{fig:trotAcceleration}. Discontinuities only arise around a contact switch; the desired accelerations are continuous otherwise. In the middle plot, the desired accelerations are shown when the update rate is restricted to an update frequency of \SI{20}{\hertz}. The updates are clearly visible because every time the feedforward trajectory is updated, the accumulated deviation from the feedforward plan is reset and a new open loop trajectory is tracked. In the bottom plot, we show the performance when the SLQ feedback policy is used. The discontinuities at updates are significantly reduced and smooth trajectories comparable to MPC with high update rate are retrieved.

These experiments show that using the feedback from the MPC can recover some of the performance lost due to lower update rates. By using a policy that is consistent with future MPC updates, the discontinuities at the updates are reduced up to the validity of the linear quadratic approximation. 

\subsection{Feedback Gains Near Inequality Constraints}
\begin{figure}[!t]
\centering
  \begin{minipage}[t]{1.0\columnwidth}
  \centering
    \includegraphics[width = \textwidth]{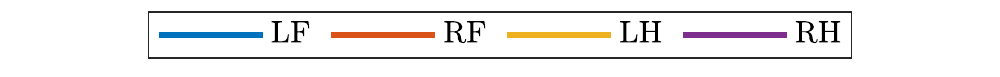}
  \end{minipage}
  \vfill
  \begin{minipage}[t]{0.48\columnwidth}
  \centering
    \includegraphics[trim=0mm 0mm 0mm 0mm, clip,width = \textwidth]{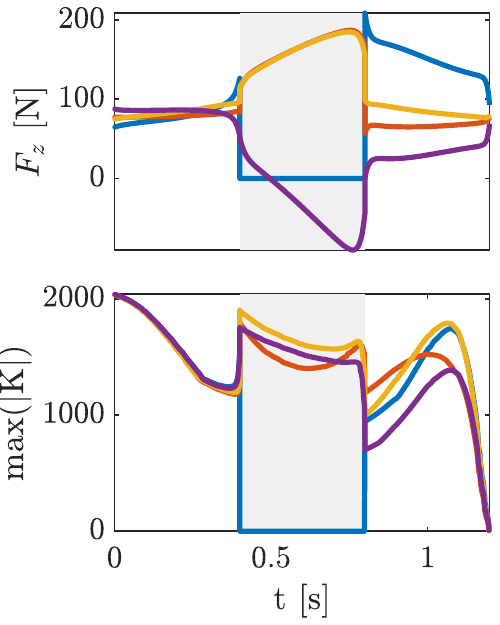}
  \end{minipage}
  \hfill
  \begin{minipage}[t]{0.48\columnwidth}
  \centering
    \includegraphics[trim=0mm 0mm 0mm 0mm, clip,width = \textwidth]{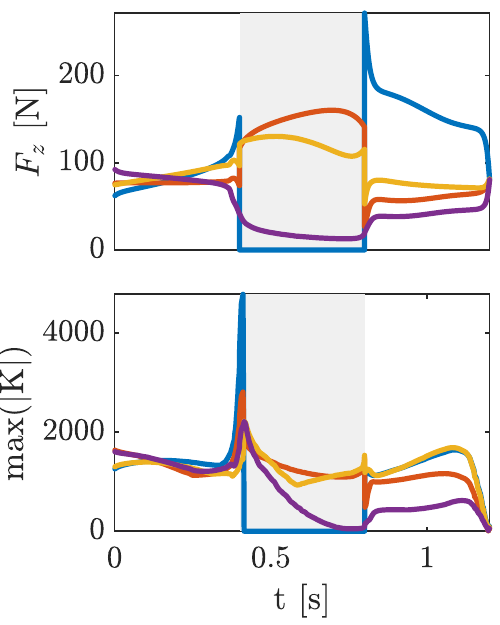}
  \end{minipage}
  \vspace{-3mm}
  \caption{Optimal vertical contact force trajectory and maximum feedback gain associated with that contact force when planning a simultaneous step and reach task without (left), and with (right) inequality constraints. The time where the left front (LF) is in the air is marked in gray.}
  \label{fig:baseline_ineq}
  \vspace{-3mm}
\end{figure}

In the following experiment we prescribe a task where we require to lift the left front leg and simultaneously set the desired body location towards the front left, outside of the support polygon.
This task requires the algorithm to coordinate the step with the body movement. The experiment is performed without the frequency shaped cost to allow us to focus on the effect of using the inequality constraints. In the left side of Fig.~\ref{fig:baseline_ineq} we show the optimized solution without inequality constraints. Without the cone constraint, the optimal strategy is to produce negative contact forces in the right hind leg, such that the desired body position can be reached as early as possible. Furthermore, we plot the maximum gain in the row associated with each vertical contact force, which shows that the zero force feedback terms for the swing leg are compliant with the corresponding equality constraint. 

The resulting solution when adding inequality constraints is shown on the right of Fig.~\ref{fig:baseline_ineq}. Here, we used fixed barrier parameters $\mu=0.5$, $\delta=0.1$, under which all constraints are strictly satisfied. With the inequality constraints, the contact forces remain positive on the right hind leg. Additionally, where before the feedback gains are about equal for the three remaining stance legs, now, the feedback gains for the right hind leg are significantly reduced. When the contact force approaches zero, the feedback gains go to zero as well, which ensures that the inequality constraint is not violated after applying the feedback policy at a disturbed state. This interaction between the inequality constraints and the feedback gains is a result of the barrier function. As the constraint boundary is approached, the Hessian of the barrier function w.r.t. the contact forces increases. This increases the input costs $\vR$ in \eqref{eq:feedbackmatrix_backwardpass}, and thus reduces the feedback gains.

This gradual decrease in feedback gains cannot be obtained with a clamping strategy, where the gains are unaffected, or with  an active set method, where the gains instantaneously decrease to zero when the constraint becomes active \cite{tassa2014control}.

As a secondary effect, we see that the feedback gains on the left front leg increase before lifting the leg. Because the right hind leg is forced to have low feedback gains in the upcoming phase, the body position before lifting the foot is of high importance, which is reflected in the gains.

In addition, we note that the distance to the constraint boundary can be regulated by choosing a different barrier scaling $\mu$. This means that using a finite $\mu$, in contrast to decreasing it to zero as done in an interior point method, can be used to trade some optimality for a larger stability margin.

\begin{figure*}[!t]
\centering
  \begin{minipage}[t]{0.357\linewidth}
  \centering
  \vspace{1.5mm} 
    \begin{minipage}[t]{1.0\linewidth}
        \centering
        \includegraphics[width = \linewidth]{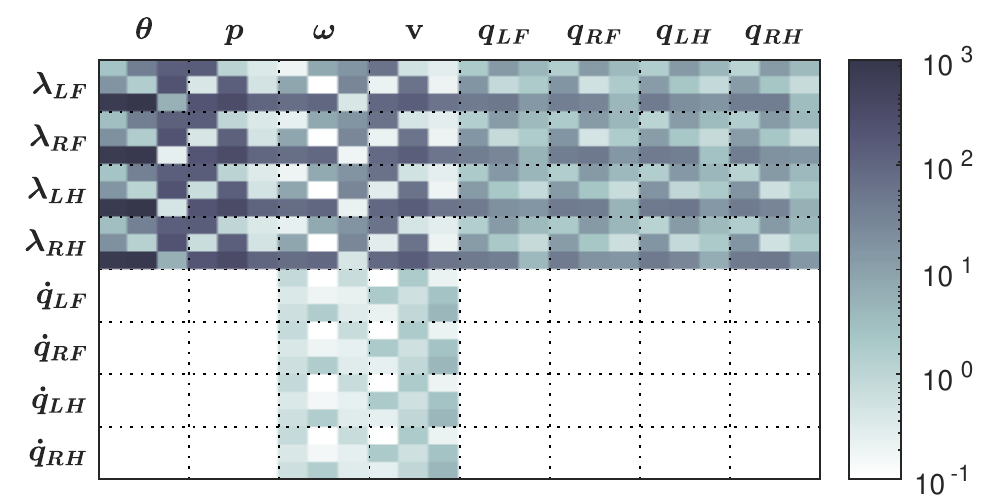}
        \footnotesize{(a)}
    \end{minipage}
    \begin{minipage}[t]{1.0\linewidth}
      \caption{The feedback matrix in stance configuration without (a) and with (b) frequency shaped cost. The color of each state-input entry represents the magnitude of the feedback. In (b), the matrix is split into four parts with input and states associated with each block marked at the left and bottom. The top left for example shows the gains between system states, $\vx$, and system inputs.}\label{fig:gains}
     \end{minipage}
  \end{minipage}
  \hfill
  \begin{minipage}[t]{0.632\linewidth}
  \centering
  \vspace{0mm}
    \includegraphics[width = \linewidth]{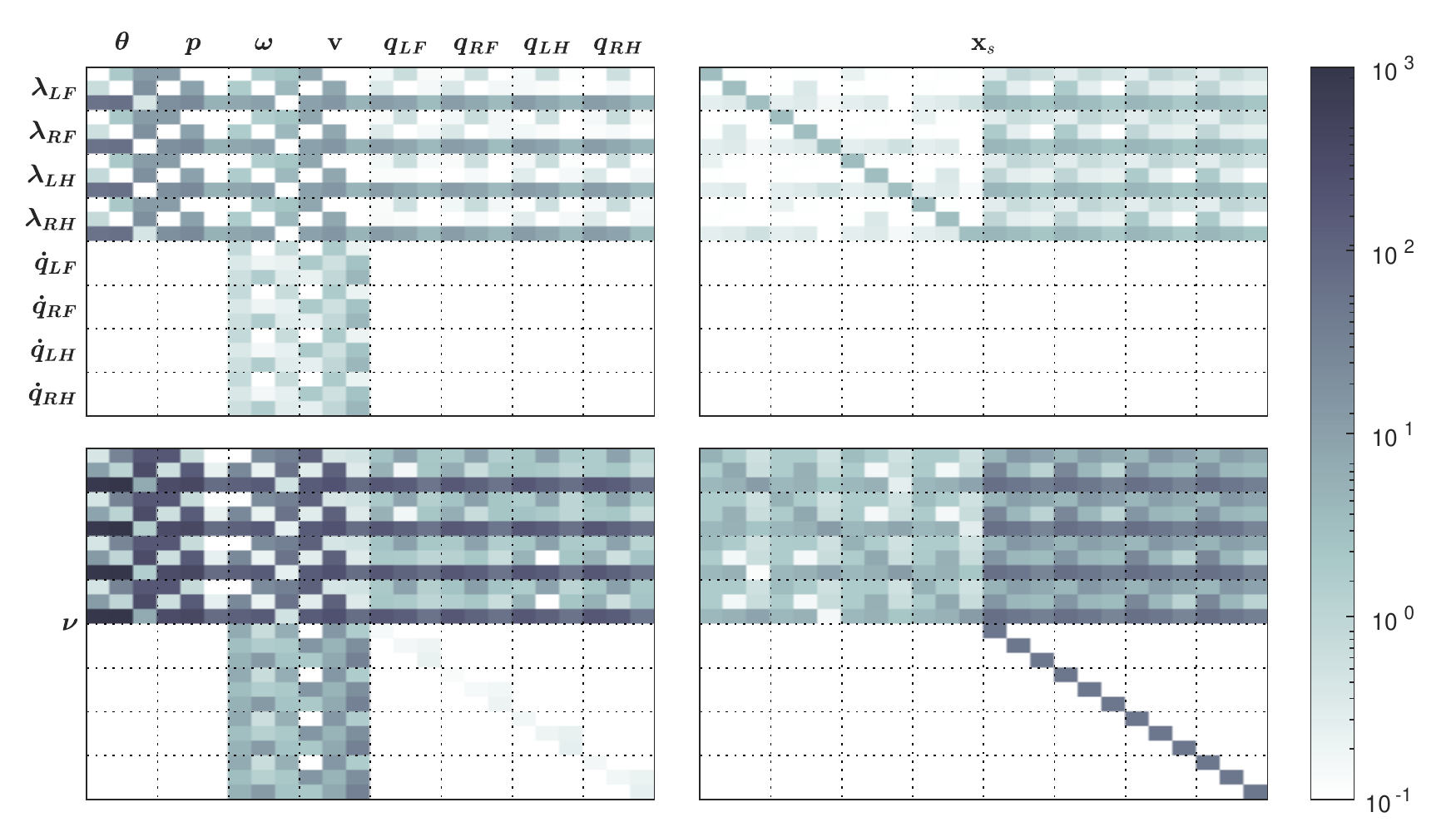}
      \vspace{-5mm}
    \footnotesize{(b)}
  \end{minipage}%
\end{figure*}

\subsection{Feedback Structure}
\label{Sect:feedback_structure}
We visualize the feedback matrix for ANYmal in a full stance configuration. The gains obtained without using the frequency-dependent cost function are shown in Fig.~\ref{fig:gains}a. The state acting on each column of the matrix is shown above the figure, and the control input affected by each row is shown on the left. The color intensity shows the magnitude of each entry in the feedback matrix, with zero shown as white and the highest gain shown in black.

In the joint velocity part of the feedback matrix, one can see how the equality constraints that require zero velocity at the end-effectors are reflected: The joint velocity commands are highly dependent on the linear and angular velocity of the base to achieve this constraint. This empirically verifies that \eqref{eq:feedbackmatrix_backwardpass} indeed produces feedback matrices that are consistent with the constraints.

The feedback matrix in Fig~\ref{fig:gains}a can be compared to the feedback matrix obtained when using the frequency shaped cost function, shown in Fig~\ref{fig:gains}b. 
We split the feedback matrix in Fig~\ref{fig:gains}b into four parts, with the vertical split between system and filter state, and horizontal split between system and auxiliary input, corresponding to the partitioning in \eqref{eq:freq_feedback}. First, the left side is inspected. Here we recognize a feedback pattern similar to that in Fig~\ref{fig:gains}a, obtained without the frequency-dependent cost function. Indeed since the shaping functions in \eqref{eq:individualweightingfunction} have unit DC-gain, the feedback matrix $K_{\vnu,\vx}$ is expected to be approximately equal to the matrix obtained without frequency shaping. Furthermore, we can recognize that the direct gains from state to contact forces are an order of magnitude smaller than before. 

For the frequency shaped feedback policy, the main feedback flows from system state to auxiliary inputs. The auxiliary inputs then drive the filter states, $\vx_s$, which in turn provide the adaptation of the system contact force inputs. 

For the joint velocities, however, the feedback from filter states to joint velocities is zero, as expected. The fact that end-effector velocities are constrained to be zero is still reflected in the feedback matrix in exactly the same way as before. Feedback terms therefore appear in the bottom right partition of Fig~\ref{fig:gains}b, to satisfy the equality constraint, $\vu = \vC_s \vx_s + \vD_s \vnu$ , for the rows associated with joint velocities.

This analysis shows that using the frequency-dependent cost function introduces smoothness and reduced direct gains where possible, but at the same time still respects hard equality constraints on the original system inputs.

\begin{figure}[!t]
    \centering
    \includegraphics[width=0.9\columnwidth]{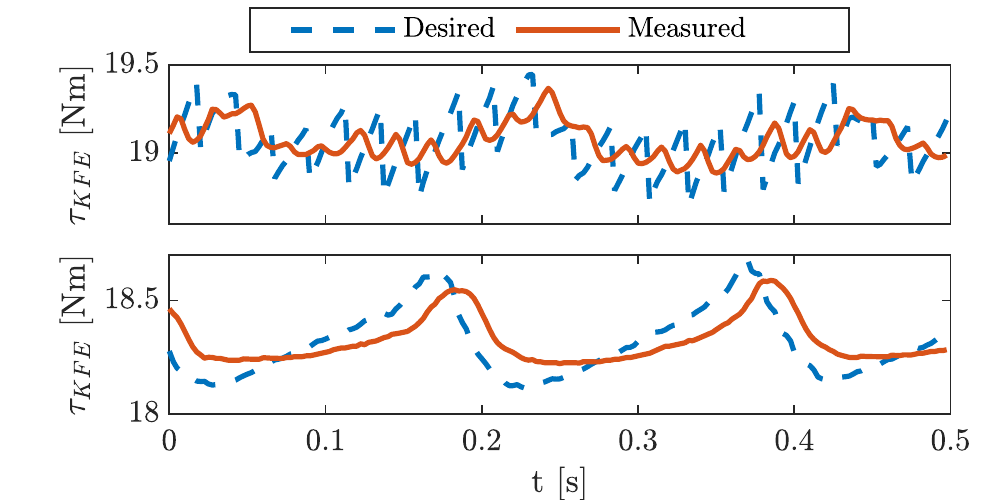}
    \vspace*{-2mm}
    \caption{Desired and measured torque signals in the left front knee under a constant disturbance of \SI{5.7}{\kilo\gram}. The desired torque is the result of both feedforward and feedback signals. Where in the top plot the feedback is provided by the conventional tracking controller, in the bottom plot the feedback from the frequency shaped SLQ algorithm is used.}
    \label{fig:disturbanceExperiment}
    \vspace{-3mm}
\end{figure}

\subsection{Hardware Experiments: Disturbance Rejection}
As seen in the accompanying video\footnote{A video of the experiments is available at \url{https://youtu.be/KrTrLGDA6FQ}}, when LQR gains obtained from the SLQ-MPC are used on hardware, the system becomes unstable even in full stance phase. As demonstrated in Section~\ref{Sect:feedback_structure}, the frequency shaped formulation reduces the direct gains between state and input, which enables successful deployment on hardware. All hardware experiments are therefore performed with both frequency shaping and inequality constraints active. 

We first perform a simple experiment to verify the qualitative difference observed in simulation between using only a feedforward policy with a conventional tracking controller or when using a feedback policy. The robot is put in a standing configuration with the desired position set to the initial position with zero velocity. Afterward, we place a \SI{5.7}{\kilo\gram} mass ($\approx 15 \%$ of the total mass) on top of the robot to induce a constant disturbance. Fig.~\ref{fig:disturbanceExperiment} shows the resulting desired and measured torque trajectories in the left front knee after the system reaches an equilibrium. 

In the top plot, the points at which the feedforward policy is updated are clearly visible. At each update, the state reference is reset to the measured reference, which effectively nullifies the feedback of the tracking controller. Since the feedforward control signal does not account for the additional disturbance, the system deviates from the desired trajectory and builds up feedback in the tracking controller until the next update arrives. In the bottom plot, where the feedback policy updates arrive at the same rate as the feedforward case, there are no discontinuous jumps in desired torque. This verifies our earlier observation in simulation.

\subsection{Hardware Experiments: Dynamic Walking}
\begin{figure}[!t]
    \centering
    \includegraphics[width=1.0\columnwidth]{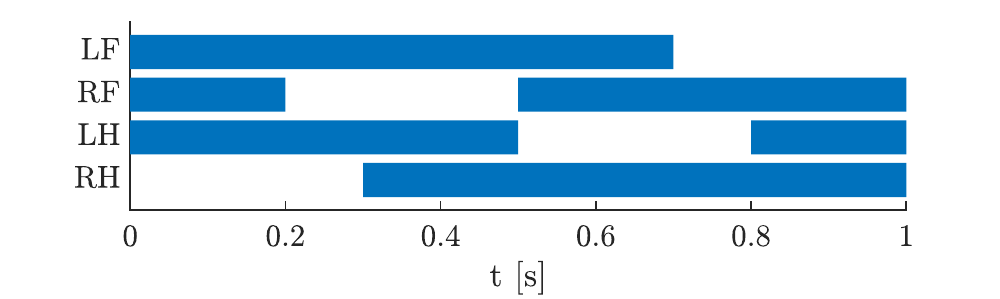}
    \vspace*{-7mm}
    \caption{Gait pattern for the dynamic walk used in the hardware experiments. Colored areas represent that a leg is in contact.}
    \label{fig:gaitPattern}
    \vspace{-3mm}
\end{figure}

Finally, we demonstrate that the proposed method achieves stable walking with all computations running on the onboard computers. We use a gait known as \textit{dynamic walk}. The gait pattern is shown in Fig.~\ref{fig:gaitPattern}.
It consists of a mixture of underactuated and overactuated contact configurations when two and three feet are on the ground, respectively.
The proposed friction cone constraint ensures that the trajectory does not require negative contact forces and thus successfully navigates the intriguing pattern of support polygons.

\begin{figure}[!t]
    \centering
    \includegraphics[width=1.0\columnwidth]{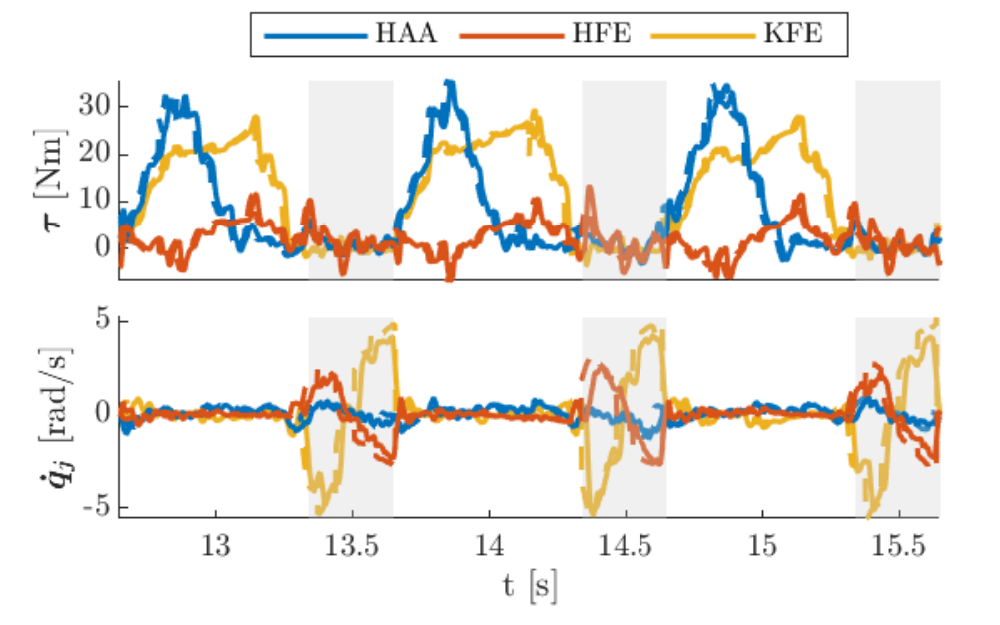}
    \vspace{-7mm}
    \caption{Torque and joint velocity trajectories during three cycles of a dynamic walk with ANYmal. The desired (dashed line) and measured (full line) are shown for the HAA (Hip Abduction Adduction), HFE (Hip Flexion Extension), and KFE (Knee Flexion Extension).}
    \label{fig:dynamic_walk}
    \vspace{-3mm}
\end{figure}

A receding horizon of \SI{1.0}{\second} was used, for which the MPC reaches an update frequency of approximately \SI{15}{\hertz}. 
The resulting desired and measured torque and joint velocity trajectories are shown in Fig.~\ref{fig:dynamic_walk} for the left front leg. The desired and measured signals are close to each other at all time, showing that the applied feedback policy respects the bandwidth limits of the actuators. More dynamic gaits such as pace and trot motions are feasible as well. In the linked \href{https://youtu.be/KrTrLGDA6FQ}{video} a continuous transition between the two gaits is shown. The feedback MPC is able to skilfully coordinate leg and body motions in these unstable and underactuated situations. 

\section{Conclusion}
In this work, we proposed to use feedback MPC as an effective way to handle the slow update rate associated with the computational restrictions of mobile platforms. The sensitivity of CoM control to uncertainties and sampling period has been recently analyzed in \cite{Villa2019}, which provides a theoretical basis for our observation that stable walking is possible with low update rates.

We proposed a relaxed barrier function method to extend the SLQ algorithm to optimization problems with inequality constraints. In particular, the friction cone is implemented through a perturbed second-order cone constraint. This formulation adds a convex penalty to the cost function and avoids numerical ill-conditioning at the origin of the cone.

A frequency-aware MPC approach was used to systematically include the bandwidth limit of the actuators in the feedback policy design. 
This was a key factor to achieve closed-loop stability on hardware without any detuning of the low-level actuator controllers as suggested in \cite{mason2016}. The frequency-aware approach effectively allows to set high gains in the low-frequency spectrum and to attenuate gains in high frequency. 
It thus increases the robustness of the feedback policy in the presence of high-frequency disturbances. 
We showed that the feedback policy is consistent with the constraints of the locomotion task. 
We empirically confirmed that the MPC policy reduces the feedback gains near the boundaries of the friction cone to respect the inequality constraints. 
We also demonstrated that the optimized policy sets zero gains on the contact force of the swing legs and encodes the zero end-effector velocity constraint for stance legs to satisfy state-input equality constraints.

\bibliographystyle{myIEEEtran} 
\bibliography{IEEEabrv,IEEEexample,bibliography}

\end{document}